\documentclass[review]{elsarticle}
\usepackage{setspace}
\usepackage{newtxmath}
\usepackage{adjustbox}
\usepackage{array}
\usepackage[usenames,dvipsnames]{color}
\usepackage{float}
\usepackage{algorithm}
\usepackage{algpseudocode}
\usepackage{caption}
\usepackage{xr}
\usepackage[numbers,sort&compress]{natbib}

\biboptions{numbers,sort&compress}
\journal{Journal of Manufacturing Processes}

\title{Physics-informed data-driven machine health monitoring for two-photon lithography}

\author[add1]{Sixian Jia\fnref{equalcontrib}} 
\ead{sixian@umich.edu}

\author[add2]{Zhiqiao Dong\fnref{equalcontrib}} 
\ead{zhiqiao5@illinois.edu}
\fntext[equalcontrib]{These authors contributed equally to this work.}

\author[add1,add2]{Chenhui Shao\corref{mycorrespondingauthor}}
\cortext[mycorrespondingauthor]{Corresponding author: Chenhui Shao (chshao@umich.edu).}
\ead{chshao@umich.edu}

\address[add1]{Department of Mechanical Engineering, University of Michigan, Ann Arbor, MI 48109, United States}
\address[add2]{Department of Mechanical Science and Engineering, University of Illinois at Urbana-Champaign, Urbana, IL 61801, United States}

\begin{document}
\begin{abstract}

Two-photon lithography (TPL) is a sophisticated additive manufacturing technology for creating three-dimensional (3D) micro- and nano-structures. Maintaining the health of TPL systems is critical for ensuring consistent fabrication quality. Current maintenance practices often rely on experience rather than informed monitoring of machine health, resulting in either untimely maintenance that causes machine downtime and poor-quality fabrication, or unnecessary maintenance that leads to inefficiencies and avoidable downtime. To address this gap, this paper presents three methods for accurate and timely monitoring of TPL machine health. Through integrating physics-informed data-driven predictive models for structure dimensions with statistical approaches, the proposed methods are able to handle increasingly complex scenarios featuring different levels of generalizability. A comprehensive experimental dataset that encompasses six process parameter combinations and six structure dimensions under two machine health conditions was collected to evaluate the effectiveness of the proposed approaches. Across all test scenarios, the approaches are shown to achieve high accuracies, demonstrating excellent effectiveness, robustness, and generalizability. These results represent a significant step toward condition-based maintenance for TPL systems.

\end{abstract}
\maketitle

\section{Introduction}
Two-photon lithography (TPL), also known as direct laser writing, is an additive manufacturing (AM) method for fabricating micro- and nano-scale structures \cite{harinarayana2021two, vyatskikh2018additive, maddox2020digitization,dong2024filtered,sun2025emerging}. In TPL, an ultrafast laser beam with high intensity is focused within a photo-reactive polymer and the concentrated high-intensity region of the focal volume can be polymerized \cite{williams2018two,jia2025end,sun2024automated}. Three-dimension (3D) fabrication using TPL was first demonstrated in 1997 \cite{maruo1997three} and has received extensive attention since then. 3D micro- and nano-scale structures have significant applications in nano-optics \cite{shunhua2023high, beermann2008two}, metamaterial engineering \cite{noronha2024titanium,surjadi2019mechanical}, and advanced electronic technologies \cite{jalali2006raman,langfelder2010mems}, etc.

TPL is a high-precision process, in which the quality and geometric accuracy of fabricated structures are determined by multiple factors such as machine conditions, process parameters (e.g., laser power, scanning rate), and the choice of photoresist \cite{zhou2015review, wang2023two, lee2023enhanced}. Among these, machine conditions have proven to be a major source of quality issues. Routine preventive measures, such as laser path alignment \cite{yu2024two}, laser source replacement, optical train realignment and cleaning (objectives, mirrors, and windows), stage calibration, and environmental stabilization (temperature, humidity, and vibration), are essential for preserving beam quality, focal integrity, and motion accuracy. When maintenance is delayed or inadequately performed, system performance can be disrupted, leading to defects such as the step effect, misalignment, and inconsistent voxel size. These problems introduce significant geometric errors, increase material waste, and cause costly downtime, ultimately reducing both production efficiency and profitability.

However, there is very limited work on machine health monitoring for TPL. In practice, current maintenance routines are often based on operator experience, rule-of-thumb schedules, or reactive interventions after a problem has already occurred. Such approaches can be problematic: untimely maintenance may lead to unexpected machine downtime, degraded beam quality, poor structural quality, or significant geometric errors in fabricated parts, while unnecessary or overly conservative maintenance introduces inefficiencies, avoidable downtime, and additional operating costs. Existing studies on TPL quality control have focused on geometric accuracy and part quality through directly examining the fabricated parts, rather than addressing machine health. For example, Lee et al. demonstrated how structure quality is affected by dosage, which depends on scanning speed and laser intensity \cite{lee2020automated}. Yang et al. proposed a machine-learning-based framework to improve geometric compliance in TPL under the same process settings \cite{yang2022machine}. Jia et al. introduced a hybrid physics-guided, data-driven modeling framework to predict geometric accuracy across different laser parameters and structure dimensions \cite{jia2024hybrid}. More recently, Jia et al. developed an end-to-end computer vision method for classifying the quality of TPL-fabricated parts \cite{jia2025end}. Although these advances significantly enhance structural quality through parameter optimization and geometry modeling, they do not explicitly account for machine condition or maintenance events.

Despite the scarcity of research specifically on TPL machine health monitoring, macro-scale additive manufacturing has seen growing interest in leveraging theoretical analysis and machine learning approaches for monitoring machine health \cite{khanafer2024condition,zhu2021metal,fang2022process}. For instance, nozzle condition has been investigated as a key contributor to part quality degradation in extrusion-based printing systems \cite{liu2018improved}. Tlegenov et al. proposed a method using vibration sensing to monitor nozzle condition, specifically for detecting clogging in fused deposition modeling, and introduced a supporting theoretical model relating nozzle condition to the process \cite{tlegenov2018nozzle}. In laser directed energy deposition, recent work has shown that machine-learning-based monitoring and control can maintain processing stability and consistency by detecting parameter drift and adjusting process settings in situ \cite{wang2024maintenance}. However, it should be noted that these methods are not directly applicable to TPL due to the significant differences in the process mechanisms and part scales.

To address this research gap, we present three distinct machine health monitoring methods for TPL, which integrate physics-informed predictive models of structure dimensions with statistical approaches and are adaptable to various production scenarios with different levels of data availability. The first method is applicable for monitoring parts produced under identical design and process settings and utilizes a two-sample t-test, achieving up to 100\% accuracy. The second method extends monitoring to new designs by combining physics-informed models with Hotelling’s $T^2$ test, yielding a classification accuracy of 83.33\% and 86.11\% for scenarios involving different and same machine conditions, respectively. The third approach monitors the physics-informed model parameters themselves, offering more generalizability than the first two methods. Leveraging bootstrap resampling and Hotelling's $T^2$ test, this approach achieves an accuracy of 100\% for distinguishing different machine statuses. To address the more challenging scenario of monitoring parts from a previously unseen process parameter group, a novel leave-one-out thresholding algorithm with majority voting is further developed. This method achieves monitoring accuracies of 82.29\% and 83.82\% for situations involving same machine conditions and different machine conditions, respectively.

The remainder of this paper is organized as follows. Section \ref{sec:DOE} details the experimental design and data collection process. Section \ref{sec:method1}, Section \ref{sec:method2}, and Section \ref{sec:method3} present the three proposed machine health monitoring methods and evaluate their performance. Finally, Section \ref{sec:conclusion} concludes the paper and outlines potential avenues for future research.

\section{Design of Experiments, Data Collection, and Evaluation Metrics} \label{sec:DOE}

\subsection{Design of Experiments and Data Collection}

\begin{figure}[h]
\centering
\includegraphics[width= 0.85 \linewidth]{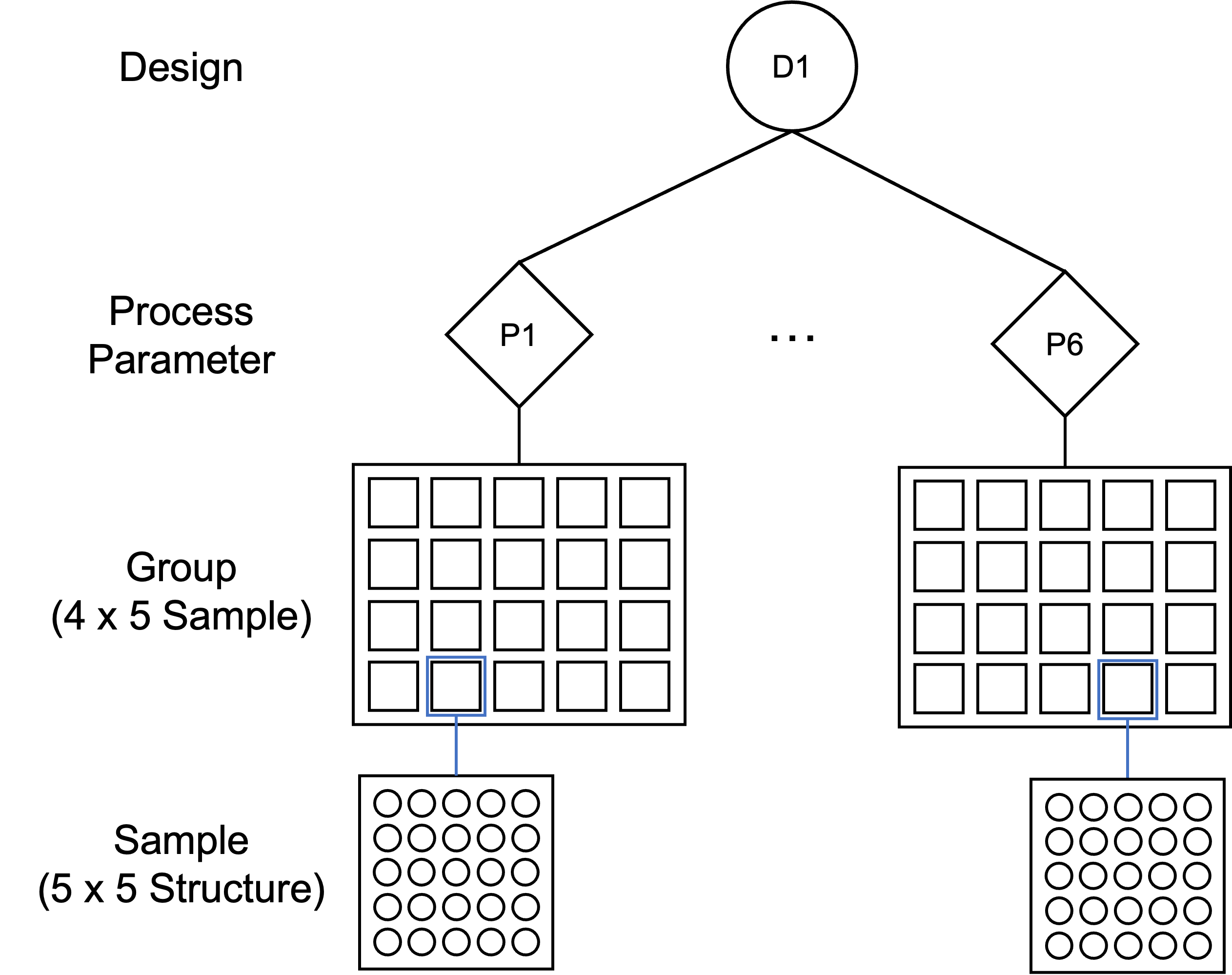}
\caption{Experiments were repeated for 
 six hemisphere radius structure design dimensions. Each design was fabricated using six different parametric combinations \cite{jia2024hybrid}.}
\label{fig:DOE}
\end{figure}

Experiments were conducted under two different machine health conditions. The design of experiments is illustrated by Fig. \ref{fig:DOE}. Each set of experiment has six different designs of hemisphere structures varied from 1.6 µm to 2.6 µm with a 0.2-µm increment. Six parametric combinations P1--P6 including the laser power (LP) and the scanning rate (SR) were used for each structure design dimension as shown in Table \ref{table:1}. These structures were produced using the Photonic Professional (Nanoscibe, GT) 3D printer, which features a pulsed femtosecond fiber laser emitting at a central wavelength of 780 nm. The default setting for the power scale is 1.0 and the reference power is 50 mW.

A 3D laser scanning microscope (Keyence VK-X1000) was used to measure the geometry of fabricated structures. Each sample's dimension measurements were stored in a 768 × 1024 matrix format. Subsequently, two key geometric features, i.e., equivalent radius ($R$) and average height ($H$) were extracted. The equivalent radius is derived by quantifying the number of pixels within the surface area. Similarly, the average height of each structure is determined by averaging the height values across the flat surface area.

\begin {table}[H]
\caption{Parametric combinations used in the experiments.}
\label{table:1}
\begin{center}
\begin{tabular}{ccc}
\hline Group & LP (\%) & SR(mm/s)\\
 \hline
 P1 & 50  & 40 \\
 P2 & 50  & 60 \\
 P3 & 55  & 60 \\
 P4 & 50  & 55 \\
 P5 & 50  & 50 \\
 P6 & 50  & 45 \\
 \hline
\end{tabular}
\end{center}
\end {table}

\subsection{Evaluation Metrics}

For each proposed method, we evaluate its effectiveness under two scenarios: (1) in-control and (2) out-of-control. In the in-control situation, the TPL machine operates under stable conditions, where only random variation exists and the Type I error (false alarm rate) is assessed. On the other hand, in the out-of-control scenario, the TPL machine operates under a different condition with systematic process shifts or disturbances. In this case, we evaluate the detection power and misdetection rate (i.e., Type II error). 

\section{Method 1: Monitoring Structure Dimensions}
\label{sec:method1}

\subsection{Approach}
The first method directly monitors the geometric dimensions of hemisphere structures to identify deviations caused by machine condition changes. The logic is straightforward that if the machine status changes, even when the same design and process parameters are used, the resulting structures may exhibit measurable differences in size or shape. To formalize this idea, we employ the two-sample \textit{t}-test, a classical statistical tool for comparing the means of two populations. This test is particularly suitable here because it provides not only a measure of central tendency difference but also quantifies the confidence level of detecting such a difference under noisy experimental conditions. For each machine status, 36 datasets are collected, corresponding to the six design dimensions (D1--D6) and six parameter groups (P1--P6). Within a given design parameter combination, structures fabricated under the same LP and SR are assumed to follow the same underlying distribution.

The hypotheses are defined as follows: the null hypothesis ($H_{0}$) states that the mean of a selected geometric feature $R$ or $H$ remains the same across two machine statuses,
\begin{equation} \label{eq:H_0}
H_{0} : \mu_1 = \mu_2,
\end{equation}
while the alternative hypothesis ($H_{a}$) suggests a statistically significant shift in the mean,
\begin{equation} \label{eq:H_a}
H_{a} : \mu_1 \neq \mu_2.
\end{equation}

The \textit{t}-statistic is then computed as
\begin{equation}
t = \frac{\bar{x}_1 - \bar{x}_2}{s_p \sqrt{\frac{1}{n_1} + \frac{1}{n_2}}},
\end{equation}
where $\bar{x}_1$ and $\bar{x}_2$ are the sample means, $s_p = \sqrt{\tfrac{s^2_1+s^2_2}{2}}$ is the pooled standard deviation, and $n_1$ and $n_2$ are the numbers of samples for each group. The use of the pooled standard deviation ensures that the variability from both machine statuses is incorporated into the comparison.


\subsection{Results and Discussion}
\label{sec:method1_results}
\begin{figure}[h]
\centering
\includegraphics[width=1\linewidth]{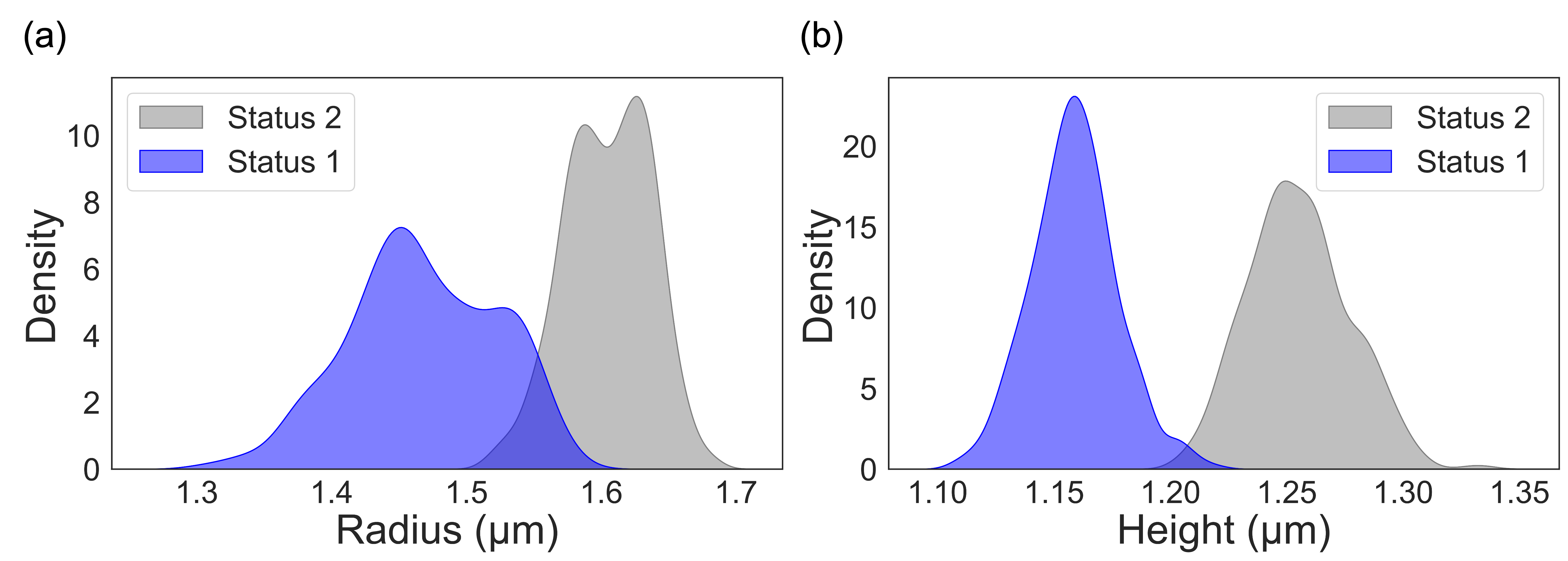}
\caption {Distribution of geometric features under two machine statuses: (a) $R$ distribution; (b) $H$ distribution for D3P3.}
\label{fig:distribution}
\end{figure}

Figure~\ref{fig:distribution} illustrates representative cases (D1P1 and D6P6), where the distributions of equivalent radius and average height clearly shift under different machine conditions. Such visual evidence reinforces the hypothesis that even small deviations in machine condition status manifest as systematic changes in structure geometry.

Quantitatively, using all 20 samples per design–parameter group and per machine status, the two-sample \textit{t}-test across all 36 design parameter groups are summarized in Table~\ref{table:Two-sample t-test}. For average height $H$, the null hypothesis was rejected in all 36 cases, yielding perfect discrimination (100\% detection power). For equivalent radius, 35 of the 36 cases show significant differences, leading to a detection power of 97.22\%. Together, these results demonstrate that geometric features are highly reliable indicators of machine health when sufficient samples are available. Because our evaluation compares two different machine statuses, this setup quantifies detection power but does not provide a direct estimate of the Type I error.

\begin{table}[h]
\caption{Two-sample \textit{t}-test accuracy for monitoring structure dimensions.}
\label{table:Two-sample t-test}
\centering
\begin{tabular}{lccc}
\hline
Feature & \# Rejections & \# Non-Rejections & Accuracy (\%) \\
\hline
Equivalent Radius & 35 & 1 & 97.22 \\
Average Height    & 36 & 0 & 100   \\
\hline
\end{tabular}
\end{table}

\begin{figure}[h]
\centering
\includegraphics[width=1\linewidth]{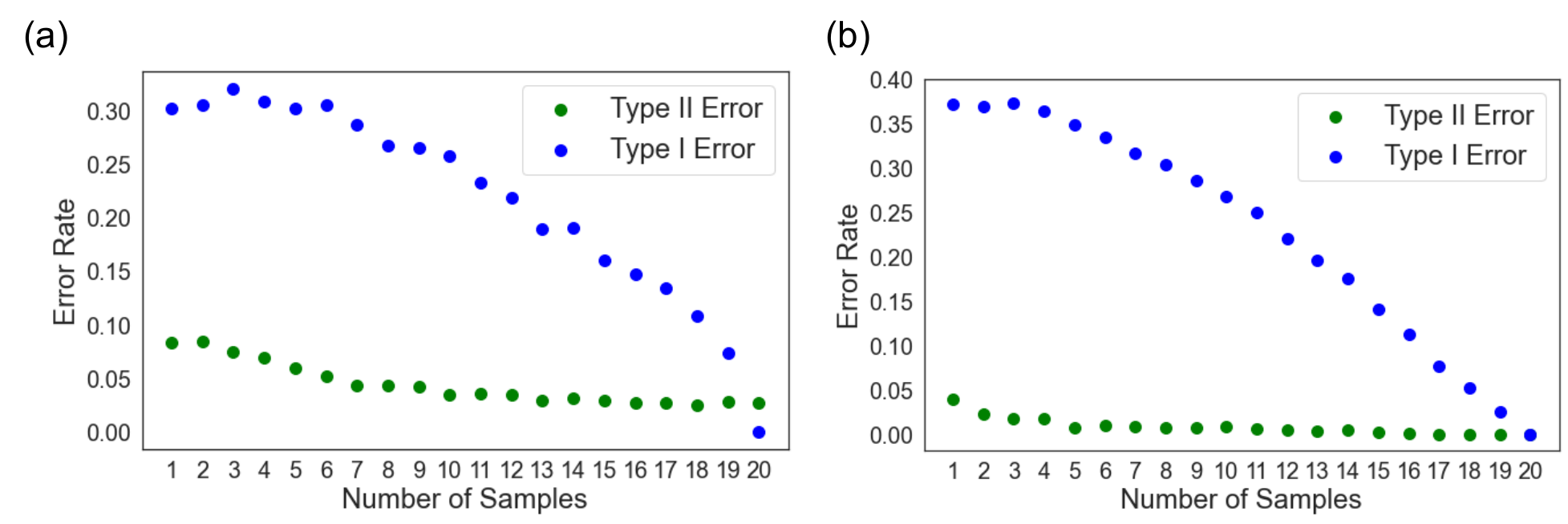}
\caption{Type I and Type II error rates of the two-sample \textit{t}-test across sample sizes: (a) radius; (b) height.}
\label{fig:data_eff}
\end{figure}

We further examine the effect of sample size on test reliability. To simulate data scarce conditions, subsets of the data were randomly drawn and the test repeated multiple times. The averaged Type I and Type II error rates are reported in Figure~\ref{fig:data_eff}. Both errors decrease as sample size increases, confirming the importance of having sufficient data to achieve robust monitoring. Notably, Type I error (false positives) is high when the number of samples is very limited, while Type II error (false negatives) decreases more gradually. This implies that while the test is highly effective in data rich scenarios, its reliability diminishes when samples are scarce.

In summary, Method 1 demonstrates that straightforward statistical monitoring of structure dimensions provides an effective and interpretable means of machine health monitoring. Its simplicity makes it an appealing choice for initial condition monitoring. However, it depends on large, paired datasets, which may not be available in some production scenarios, which highlight the need for approaches suitable for more data-limited environments.

\section{Method 2: Monitoring Structure Dimensions with New Designs}
\label{sec:method2}

\subsection{Approach}
\begin{figure}[h]
\centering
\includegraphics[width=0.6\linewidth]{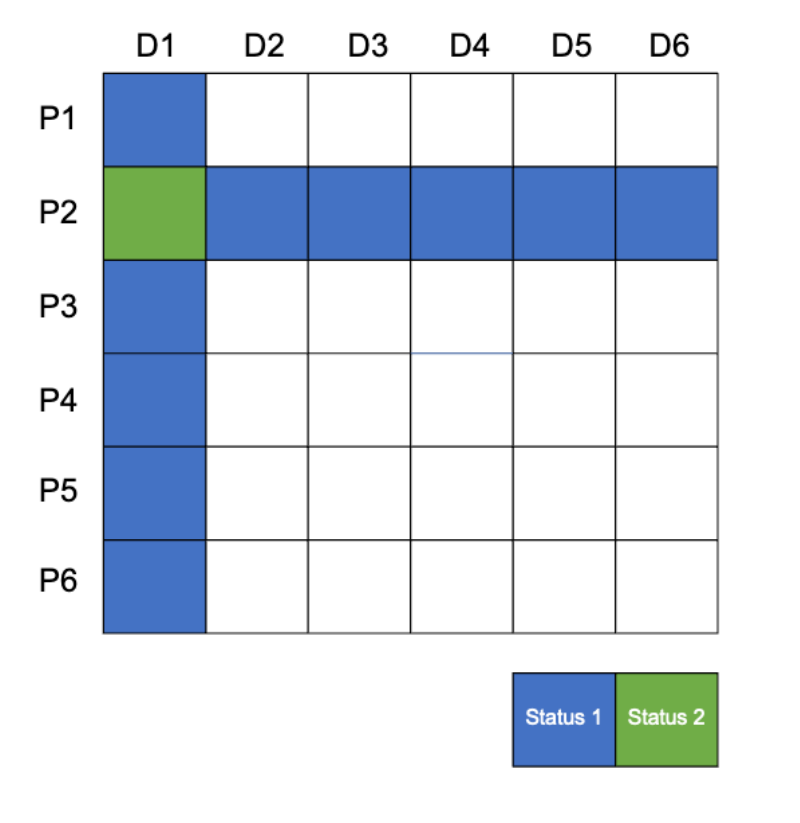}
\caption{Illustration of predicting the mean value of D1P2 for Machine Status 1 using available Status 1 datasets (blue regions).}
\label{fig:train_illustration}
\end{figure}

As shown in Section \ref{sec:method1_results}, a limitation of Method 1 is its requirement for structures produced under an identical design and parameter (DP) group for comparison. This condition restricts its general applicability in situations where corresponding datasets between machine statuses are unavailable. Figure~\ref{fig:train_illustration} illustrates this situation. Data for Machine Status 1 is available in the blue regions, while Status 2 data is only available in the green region. Without overlapping DP combinations, direct application of Method 1 is infeasible. To overcome this limitation, this section presents Method 2, which leverages physics-informed modeling to predict the expected geometric features for a given design and parameter set under a baseline machine condition.

Our prior research showed that TPL process parameters influence voxel dimensions, which subsequently affect structure dimensions~\cite{jia2024hybrid}. The predictive models for $R$ and $H$ are expressed as:
\begin{equation} \label{eq:R_model}
R(\mathrm{LP},\mathrm{SR}) = a_R \left[\ln\left(b_R \frac{\mathrm{LP}^2}{\mathrm{SR}}\right)\right]^{1/2} + c_R,
\end{equation}
\begin{equation} \label{eq:H_model}
H(\mathrm{LP},\mathrm{SR}) = a_H \left[\left(b_H \frac{\mathrm{LP}^2}{\mathrm{SR}}\right)^{1/2} - 1\right]^{1/2} + c_H,
\end{equation}
where $a(\cdot)$, $b(\cdot)$, and $c(\cdot)$ are model parameters estimated using machine learning.

By training these models with Status 1 data, we can predict the mean values of $R$ and $H$ for new DP groups, which serve as the baseline reference for comparison against Status 2 measurements. Then, the same hypotheses, shown by Equations (\ref{eq:H_0}) and (\ref{eq:H_a}), can be applied to detect machine condition changes. Here, since the statistical distributions are obtained from the unknown status and the predicted dimension values are calculated using Equations (\ref{eq:R_model}) and (\ref{eq:H_model}), we should use a one-sample $Z$-test. The $Z$-test statistic is defined as:
\begin{equation}
Z = \frac{\overline{X} - \mu_0}{s},
\end{equation}
where $\overline{X}$ is the observed sample mean from Status 2, $\mu_0$ is the predicted mean , and $s$ is the standard deviation of the sample. If $|Z|$ exceeds a chosen threshold, the null hypothesis is rejected, indicating a machine condition change.

When there is correlation between $R$ and $H$, we can also use the Hotelling’s $T^2$ test, which accounts for covariance between variables. The hypotheses are defined as follows: the null hypothesis ($H_{0}$) states that the observed mean vector equals the predicted mean vector,
\begin{equation} \label{eq:H0_T2}
H_{0} :  \boldsymbol{{\mu}_1} =  \boldsymbol{{\mu}_2},
\end{equation}
while the alternative hypothesis ($H_{a}$) suggests a statistically significant difference,
\begin{equation} \label{eq:Ha_T2}
H_{a} :  \boldsymbol{{\mu}_1} \neq  \boldsymbol{{\mu}_2}.
\end{equation}

The one-sample Hotelling’s $T^2$ statistic is then defined as:
\begin{equation}
T^2 = n(\mathbf{\overline{x}} - \boldsymbol{\mu}_0)^{\intercal} \mathbf{S}^{-1} (\mathbf{\overline{x}} - \boldsymbol{\mu}_0),
\end{equation}
where $n$ is the sample size, $\mathbf{\overline{x}}=\left[ \overline{R}, \overline{H} \right]^{\intercal}$ is the observed sample mean vector, $\boldsymbol{\mu}_0$ is the predicted mean vector following Equations (\ref{eq:R_model}) and (\ref{eq:H_model}), and $\mathbf{S}^{-1}$ is the inverse of the sample covariance matrix.

\subsection{Results and Discussion}

\subsubsection{Univariate $Z$-test}
\label{subsec:z_test}
\begin{figure}[h]
\centering
\includegraphics[width=\linewidth]{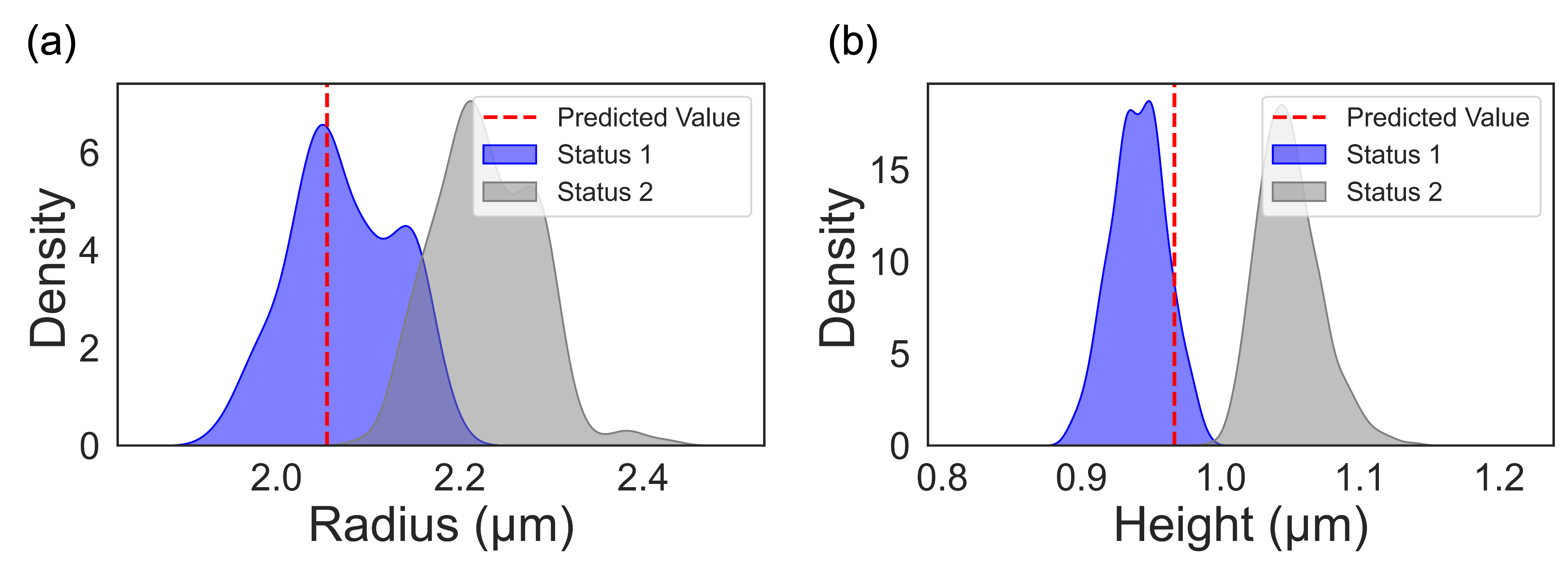}
\caption{Predicted mean values (red dashed lines) compared with observed distributions from Status 1 (blue) and Status 2 (gray): (a) Radius (D6P5); (b) Height (D2P2).}
\label{fig:prediction_result}
\end{figure}

Figure~\ref{fig:prediction_result} compares predicted means in red dashed lines with observed feature distributions from two machine statuses. For both radius and height, the predicted values align closely with Status 1 distributions but clearly deviate from Status 2, confirming that the approach captures discriminative differences between machine conditions.

Table~\ref{table:One-sample z-test} summarizes the error rates. The equivalent radius has a Type II error of 44.44\%, while average height has a Type I error of 25\%. These relatively high errors indicate that univariate $Z$-tests are not sufficient for reliable monitoring.

\begin{table}[h]
\caption{One-sample $Z$-test error rates.}
\label{table:One-sample z-test}
\centering
\begin{tabular}{lcc}
\hline
Feature & Type I Error & Type II Error \\
\hline
Equivalent Radius & 2.78\% & 44.44\% \\
Average Height    & 25\%   & 8.33\%  \\
\hline
\end{tabular}
\end{table}

\subsubsection{Hotelling’s $T^2$ test}

\begin{figure}[h]
\centering
\includegraphics[width=0.65\linewidth]{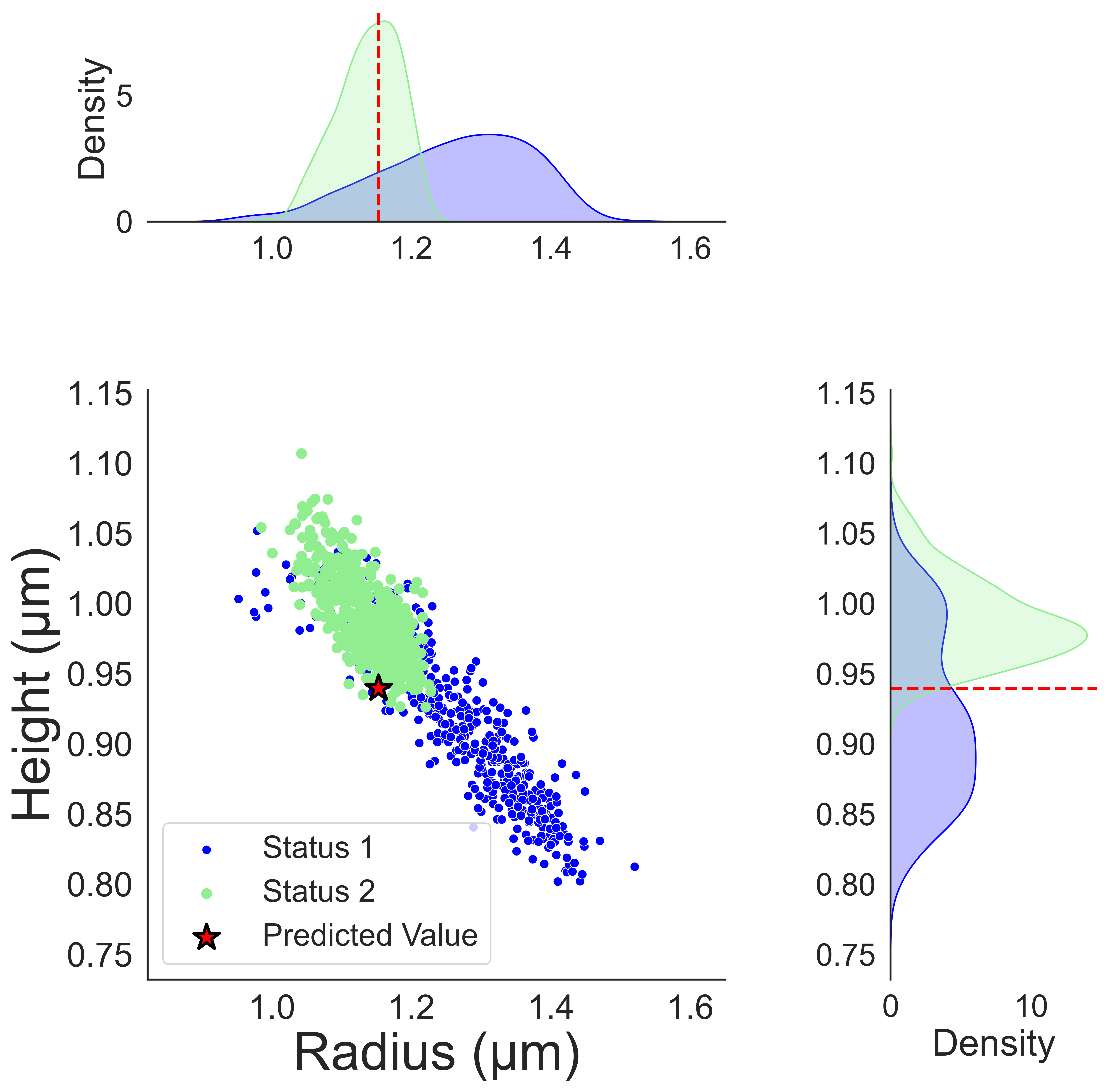}
\caption{Distribution of geometric features under two machine statuses (D1P6) with predicted values from Status 1.}
\label{fig:z_test_failed}
\end{figure}

Radius and height are actually strongly correlated with correlation coefficients of 0.94 for Status 1 and 0.95 for Status 2.. Consequently, Hotelling’s $T^2$ test, which accounts for such correlations, may outperform the univariate $Z$ tests. Figure \ref{fig:z_test_failed} illustrates an out-of-control example. In this case, the machine operated under ten DP combinations including $\text{D}_i\text{P}_6$, where $i=1,2,3,4,5$ and $\text{D}_1\text{P}_j$, where $j=2,3,4,5,6$. The goal is to test whether the machine status is changed when operating under D1P6. As shown in Figure \ref{fig:z_test_failed}, the proposed approach is able to predict the $R$ and $H$ values for D1P6 under Machine Status 1. The blue dots represent the actual measurements of $R$ and $H$ for D1P6 under Status 1; however, such data are unavailable in the test case. In Figure \ref{fig:z_test_failed}, the $Z$-test for $R$ fails to detect this change. It should also be noted that in other cases, the $Z$-test for $H$ fails, making it challenging to identify which $Z$-test could provide reliable monitoring of machine health.

\begin{figure}[h]
\centering
\includegraphics[width=1\linewidth]{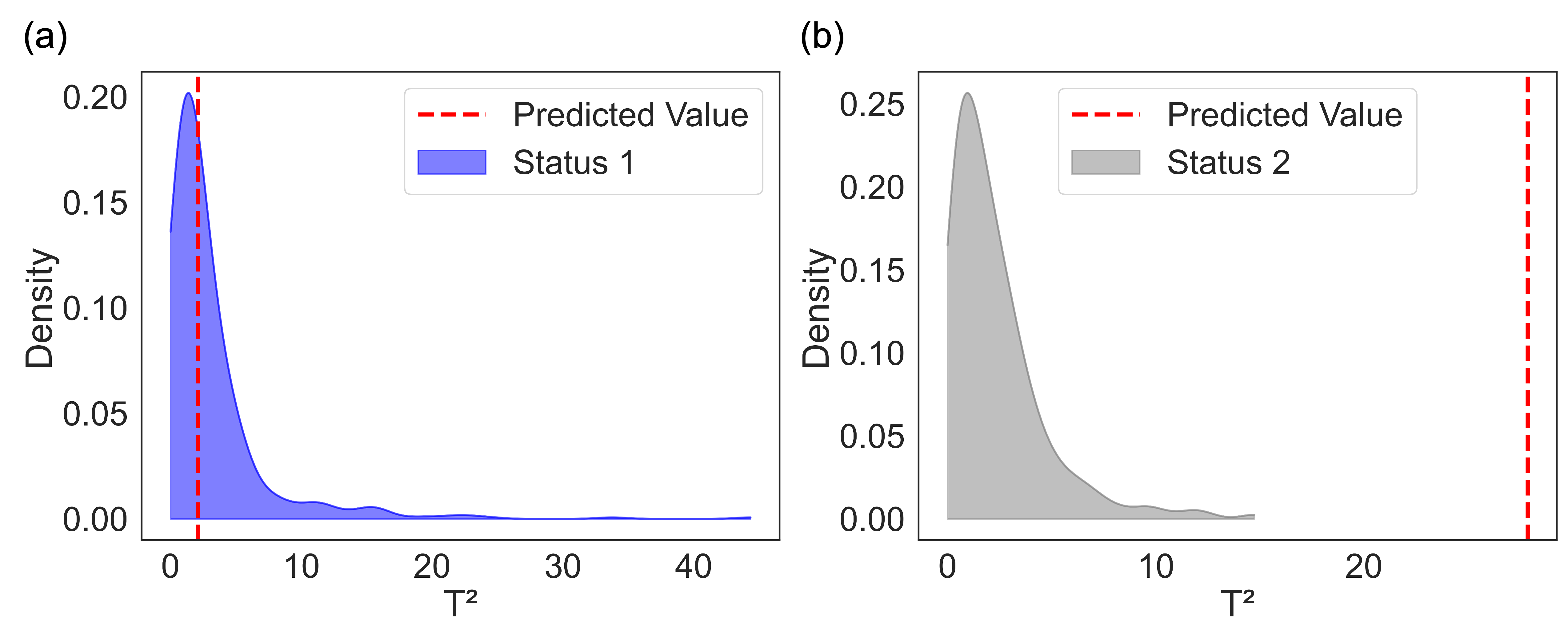}
\caption{Distribution of $T^2$  for D1P6 with predicted value: (a) Status 1; (b) Status 2}
\label{fig:T^2_distribution}
\end{figure}

In contrast, Hotelling’s $T^2$ test effectively resolves this issue and clearly indicates that the predicted value corresponds to Status 1. Figure~\ref{fig:T^2_distribution} shows the $T^2$ distributions for Statuses 1 and 2. The predicted value falls within the distribution of Status 1 but lies outside that of Status 2, thereby leading to a correct classification.

\begin{figure}[h]
\centering
\includegraphics[width=\linewidth]{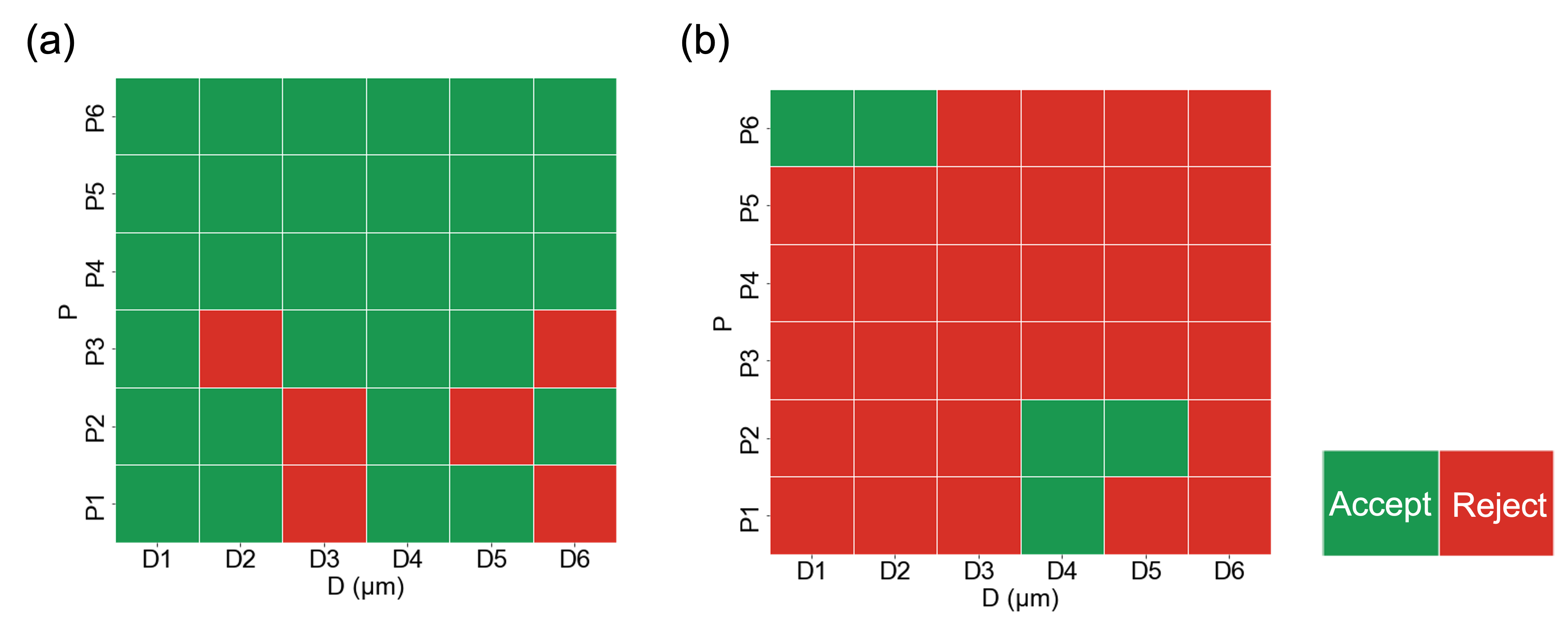}
\caption{Hotelling’s $T^2$ monitoring results for (a) in-control and (b) out-of-control scenarios. Green cells indicate non-rejection of the null hypothesis, implying that the machine status remains unchanged; red cells indicate rejection suggesting a changed machine status.}
\label{fig:hotelling_results}
\end{figure}

Results for the Hotelling’s $T^2$ test are displayed in Figure~\ref{fig:hotelling_results} and Table~\ref{table:hotelling_t2}. Compared to the $Z$ test results reported in Section \ref{subsec:z_test}, significant improvements are achieved. The accuracies for the in-control and out-of-control cases are 83.33\% and 86.11\%, respectively. This demonstrates that combining features in a multivariate test substantially enhances discriminative power.

\begin{table}[h]
\caption{Hotelling’s $T^2$ test results at 90\% confidence level.}
\label{table:hotelling_t2}
\centering
\begin{tabular}{lccc}
\hline
Status Comparison & \# Rejections & \# Acceptances & Accuracy (\%) \\
\hline
Same Status      & 6  & 30 & 83.33 \\
Different Status & 31 & 5  & 86.11 \\
\hline
\end{tabular}
\end{table}

\begin{figure}[h]
\centering
\includegraphics[width=\linewidth]{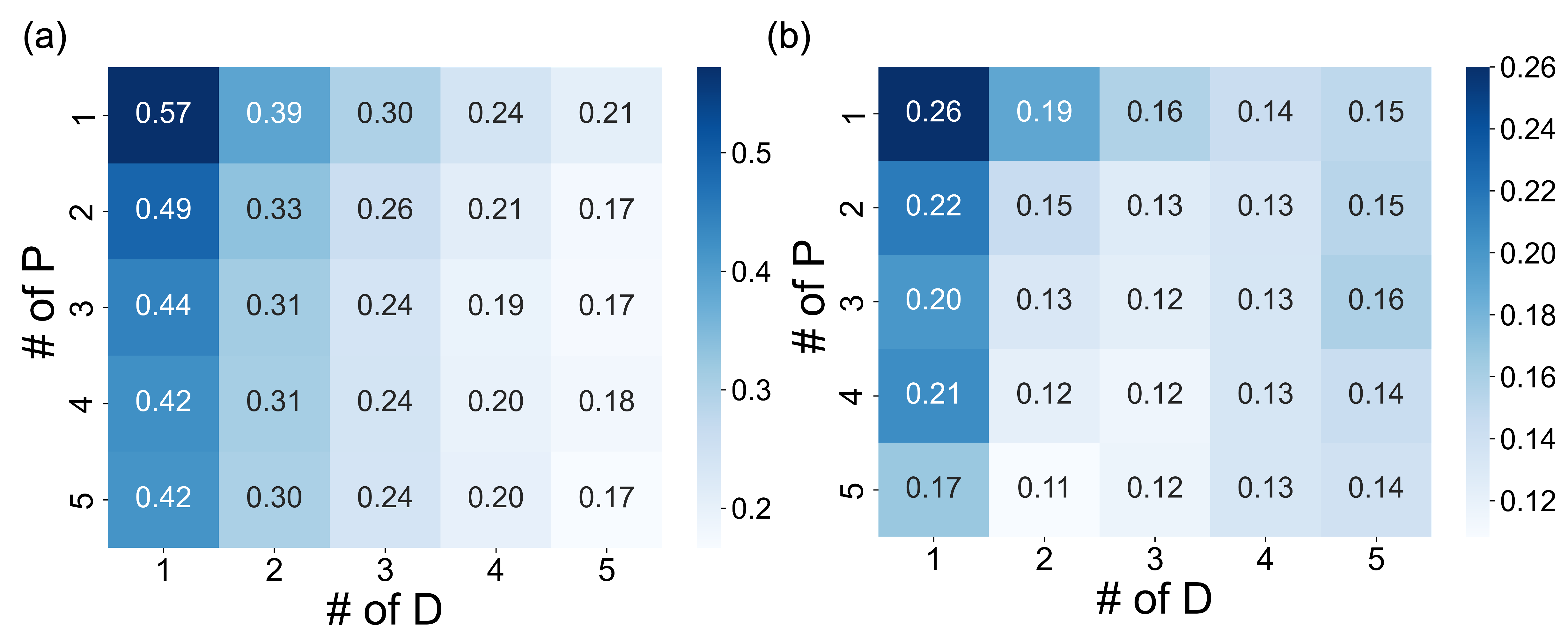}
\caption{Data efficiency of Hotelling’s $T^2$: (a) Type I Error; (b) Type II Error, as a function of the number of designs (D) and parameters (P) used for training.}
\label{fig:data_efficiency}
\end{figure}

Finally, we investigate how training set diversity affects performance. Figure~\ref{fig:data_efficiency} shows that increasing the number of designs reduces Type I errors, while increasing the number of parameter groups reduces Type II errors. This suggests that broad design coverage minimizes false positives, while parameter coverage improves sensitivity to true differences.

Method 2 extends monitoring to cases without overlapping DP combinations by combining physics-guided feature prediction with Hotelling’s $T^2$ test. The approach achieved accuracies between 83\% and 86\%, and the data efficiency analysis highlights practical guidelines: prioritize design diversity to avoid false alarms and ensure parameter diversity to reduce missed detections. This method therefore provides a more flexible and generalizable tool for TPL machine health monitoring compared to Method 1.

\section{Method 3: Monitoring Model Parameters}
\label{sec:method3}

\subsection{Approach Overview}
As detailed in Method 2, we leverage the established radius and height models (i.e., Equations (\ref{eq:R_model}) and (\ref{eq:H_model})) to predict feature values for unknown design and parameter combinations. Since the machine status influences the structure dimensions of produced parts, and those dimensions in turn determine the learned parameters of the models, an alternative strategy is to monitor the parameters themselves. Shifts in parameter values can serve as indicators of a change in machine status.

\begin{figure}[H]
\centering
\includegraphics[width=0.9\linewidth]{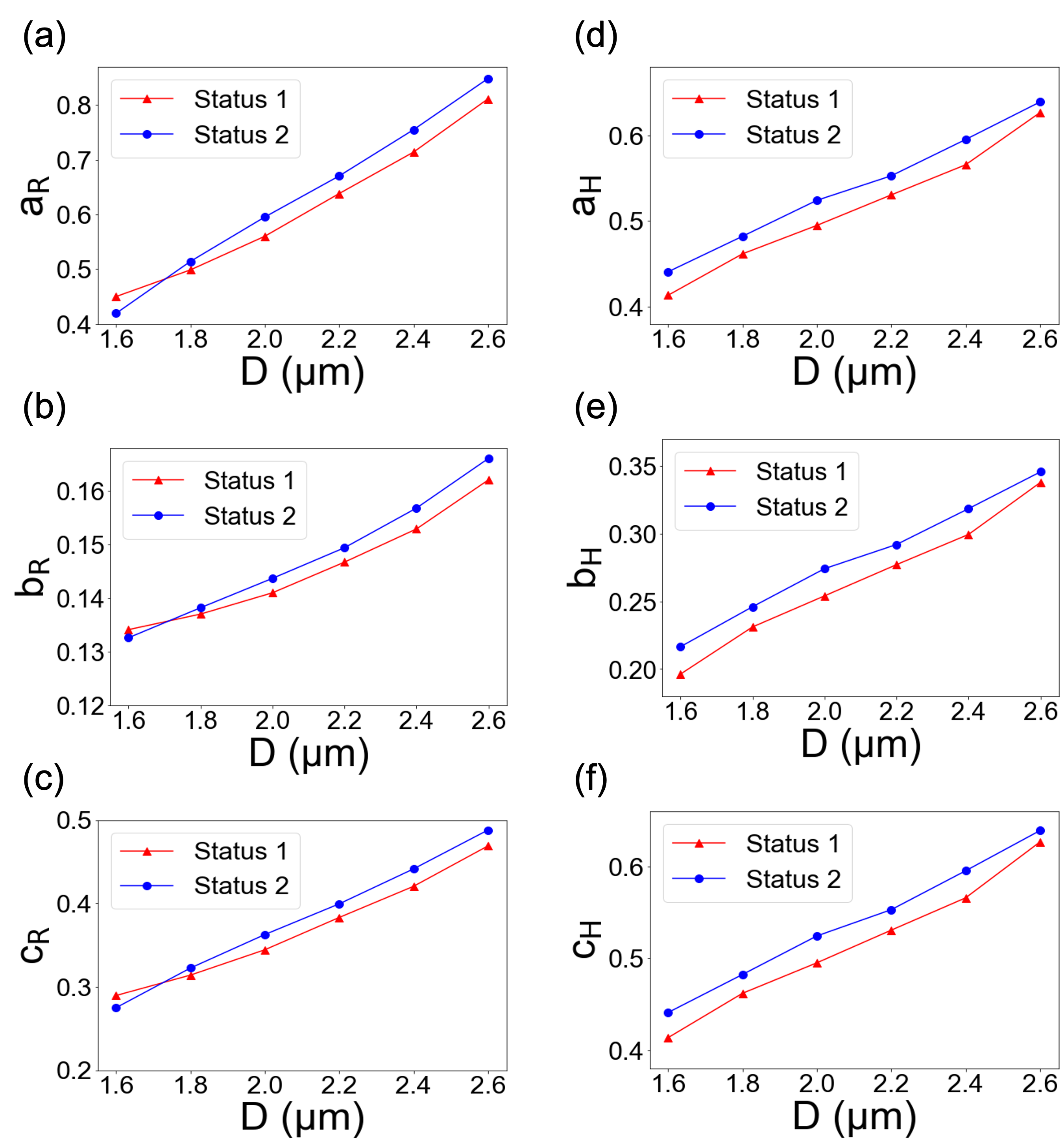}
\caption{Variation of model parameters with respect to structure dimension $D$ for Machine Status 1 and Machine Status 2: (a) $a_R$, (b) $b_R$, (c) $c_R$, (d) $a_H$, (e) $b_H$, (f) $c_H$.}
\label{fig:model_parameters}
\end{figure}

Figure~\ref{fig:model_parameters} illustrates the variation of learned model parameters, namely $a_R$, $b_R$, and $c_R$ for the radius model, and $a_H$, $b_H$, and $c_H$ for the height model, as functions of representative structure dimensions under both machine statuses. Distinct trends are visible between Status 1 and Status 2. For the radius model, parameters such as $a_R$ show divergence at specific structure dimensions even though there may be overlap in absolute values across the range. For the height model, the parameters $a_H$, $b_H$, and $c_H$ form two nearly parallel lines separated by a consistent shift, indicating systematic differences in how process parameters affect structure height under different machine conditions. This clear separation strongly suggests that model parameters themselves can provide robust indicators of machine status, even when direct measurements of radius or height may overlap between conditions.

Building on this observation, we develop two approaches to address distinct data-scarcity scenarios. In the first scenario, data for the same parameter groups are available. In this case, Method 1 can be applied; however, as shown in Figure~\ref{fig:data_eff}, monitoring errors are high when data availability is limited. The second scenario addresses a more challenging situation in which unknown parameter groups are present in the test case. Sections~\ref{subsec:same_parameter} and~\ref{subsec:unknown_parameter} describe the proposed methods and present their corresponding results, respectively.

\subsection{Monitoring Model Parameters with Same Parameter Group} \label{subsec:same_parameter}

To evaluate the stability and variability of estimated model parameters, we employ a bootstrap resampling approach. For a given design, measurement data from three parameter groups are selected, and three samples from each group are used for training. A bootstrap procedure with 40 iterations generate parameter distributions. In each iteration, data points are resampled with replacement, the geometric model is retrained, and the resulting parameters $a_R$, $b_R$, $c_R$ or $a_H$, $b_H$, $c_H$ are recorded.

The model parameters generally show a linear increasing trend with design dimension. Since each model produces three parameters, Hotelling’s $T^2$ test is used to assess whether the parameter distributions differ significantly between machine statuses. This multivariate test accounts for correlations among parameters.

\begin{figure}[h]
\centering
\includegraphics[width=\linewidth]{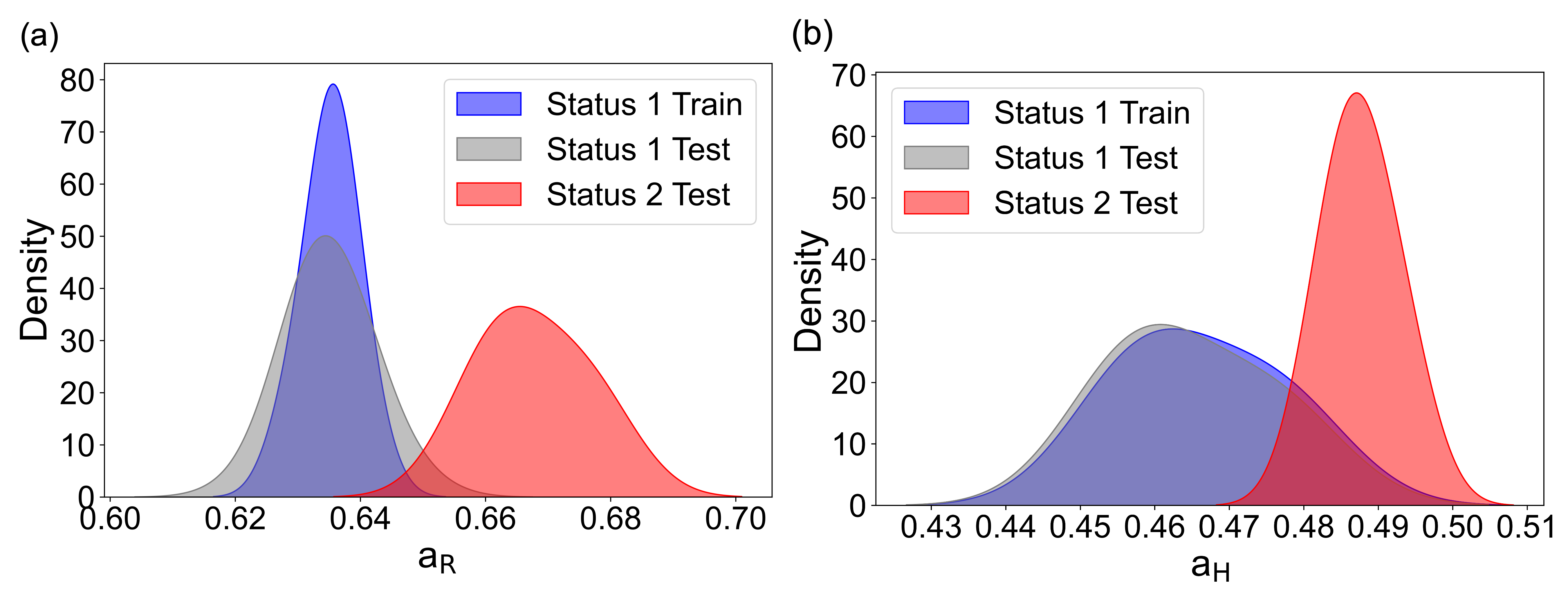}
\caption{Bootstrap-derived distributions of model parameters: (a) $a_R$ parameter for design D4 trained on P2, P3, and P6; (b) $a_H$ parameter for design D2 trained on P1, P2, and P3.}
\label{fig:parameter_distributions}
\end{figure}

Figure~\ref{fig:parameter_distributions} presents examples of bootstrap-derived parameter distributions. The distribution of $a_R$ obtained from Status 1 training data closely aligns with that from the unseen Status 1 test data, whereas the Status 2 distribution exhibits a noticeable mean shift. A similar trend is observed for $a_H$: the Status 1 distributions largely overlap, while the Status 2 distribution deviates distinctly. For brevity, the distributions of other parameters are not shown, but they exhibit consistent patterns. These observations demonstrate the sensitivity of the parameters to changes in machine condition.

The Hotelling’s $T^2$ test achieve high accuracy. For the radius model parameters, the test yield 95.83\% and 100\% accuracies for the in-control and out-of-control scenarios. Detailed results are provided in Table~\ref{table:hotelling_radius}. For the height model parameters, the corresponding accuracies are 92.50\% and 100\% as shown in Table~\ref{table:hotelling_height}.

\begin{table}[H]
\caption{Hotelling’s $T^2$ test results for radius model parameters.}
\label{table:hotelling_radius}
\centering
\begin{tabular}{lccc}
\hline
Status Comparison & \# Rejections & \# Acceptances & Accuracy (\%) \\
\hline
In-control     & 5   & 115 & 95.83 \\
Out-of-control & 120 & 0   & 100   \\
\hline
\end{tabular}
\end{table}

\begin{table}[H]
\caption{Hotelling’s $T^2$ test results for height model parameters.}
\label{table:hotelling_height}
\centering
\begin{tabular}{lccc}
\hline
Status Comparison & \# Rejections & \# Acceptances & Accuracy (\%) \\
\hline
In-control      & 9   & 111 & 92.50 \\
Out-of-control & 120 & 0   & 100   \\
\hline
\end{tabular}
\end{table}

\subsection{Monitoring Model Parameters with Unknown Parameter Group} \label{subsec:unknown_parameter}

\begin{figure}[h]
\centering
\includegraphics[width=0.55\linewidth]{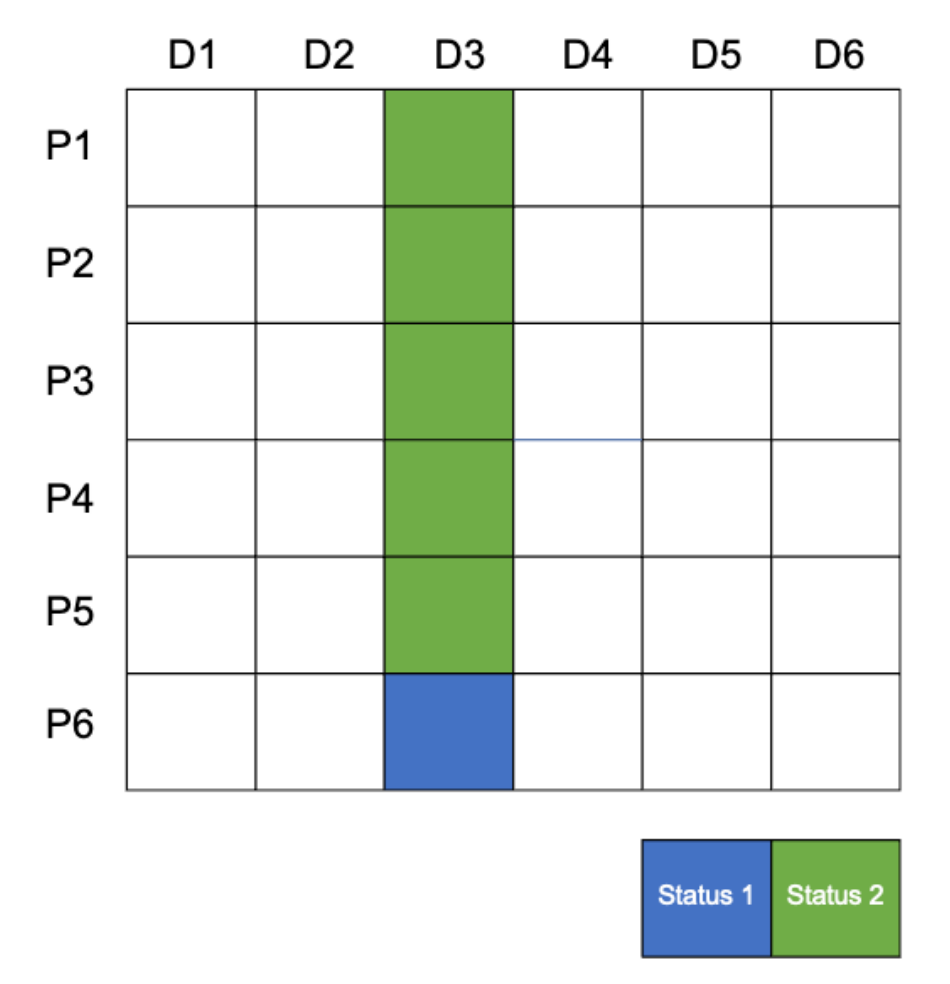}
\caption{Data usage for leave-one-out threshold estimation. Thresholds for an unseen parameter group (e.g., P6) are derived by training on subsets of P1–P5 and evaluating on the held-out group.}
\label{fig:unknown_p_data}
\end{figure}

We further address the case where new observations in testing belong to an unknown process parameter group. To do so, we develop a leave-one-out threshold estimation algorithm. The algorithm iteratively learns expected parameter ranges from subsets of known groups and evaluates coverage on the held-out group, ultimately estimating thresholds for unseen groups indicated in Figure~\ref{fig:unknown_p_data}. Algorithm~\ref{alg:threshold_unknown_p} details the procedure.

\begin{algorithm}[ht]
\floatname{algorithm}{Algorithm}
\caption{Leave-One-Out Threshold Estimation for Unknown Parameter Group}
\label{alg:threshold_unknown_p}
\begin{algorithmic}[1]
\Statex \textbf{Input:} Known process parameter groups $P = \{P_1, P_2, P_3, P_4, P_5\}$
\Statex \textbf{Target:} Unknown process parameter group $P_6$
\Statex \textbf{Output:} Estimated threshold interval $[LL_6, UL_6]$
\For {$i = 1, \ldots, 5$}
    \State Define $P_i$ as the evaluation set
    \State Construct distribution $D_i$ by bootstrapping model parameters from the remaining groups
    \State Estimate threshold interval $[LL_i, UL_i]$ using $D_i$ and $P_i$
\EndFor
\State Combine intervals $\{[LL_1, UL_1], [LL_2, UL_2], ..., [LL_5, UL_5]\}$ 
\State Output final threshold $[LL_6, UL_6]$
\end{algorithmic}
\end{algorithm}

\begin{figure}[H]
\centering
\includegraphics[width=\linewidth]{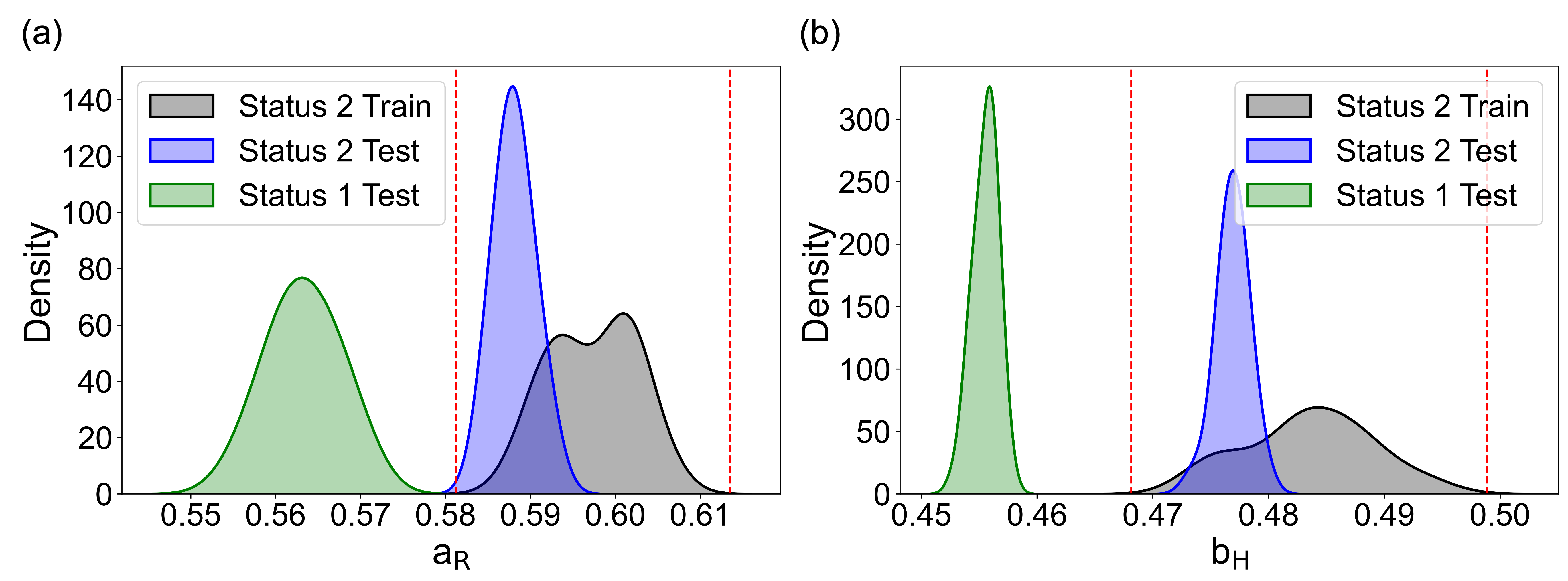}
\caption{Threshold for unknown parameter groups: (a) $a_R$; (b) $b_H$. Distributions from the same status fall within thresholds, while those from a different status fall outside.}
\label{fig:leave-one-out_res}
\end{figure}

Thresholds estimated with this approach successfully contain parameter distributions from the same machine status while excluding distributions from a different status, as shown in Figure~\ref{fig:leave-one-out_res}. This demonstrates that leave-one-out thresholding can effectively extend parameter monitoring to unseen groups.

Subsequently, hypothesis tests are conducted for all six model parameters, and majority voting is employed for machine health monitoring. Specifically, the machine is considered to have an unchanged health status if no more than two hypothesis tests are rejected; otherwise, a change in machine health is detected. The corresponding test results are summarized in Table~\ref{table:unknown_p_results}. The proposed method achieves accuracies of 82.29\% and 83.82\% for the in-control and out-of-control test scenarios, respectively.

\begin{table}[h]
\caption{Classification accuracy using majority voting across six model parameters.}
\label{table:unknown_p_results}
\centering
\begin{tabular}{lccc}
\hline
Status & \# Rejection & \# Acceptance & Accuracy (\%) \\
\hline
Same Status      & 255  & 1185 & 82.29 \\
Different Status & 1207 & 233  & 83.82 \\
\hline
\end{tabular}
\end{table}

\section{Conclusion and Future Work} \label{sec:conclusion}

This paper presented three methods for monitoring TPL machine health conditions. By integrating physics-informed models of structure dimensions with statistical methods, including the two-sample $t$-test, Hotelling's $T^2$ test, and bootstrap-based parameter monitoring, we demonstrated robust and accurate approaches for machine health monitoring. Notably, all three methods exhibited high efficacy, achieving monitoring accuracies ranging from 82.29\% to 100\% across diverse scenarios of data availability and process configurations. The successful application of these methods to a novel, large-scale experimental dataset demonstrated their potential for practical implementation in TPL manufacturing. To the best of our knowledge, this work is the first to systematically investigate condition monitoring in TPL and represents a significant step toward data-driven condition monitoring and predictive maintenance in this field.

The findings of in this study suggest several future research directions. First, the proposed approaches could be extended to monitor machine health across a wider variety of complex 3D structure designs. Incorporating transfer learning and domain adaptation techniques \cite{meng2023explainable,meng2024meta,dong2025new} may further enhance model adaptability by capturing geometric similarities and variations across diverse structures. Second, further research can focus on validating the generalizability of these methods on a larger scale and across different TPL systems and operational environments. Third, the development of real-time monitoring systems \cite{sun2023situ,sun2024automated} based on these methods could provide significant advancements in proactive maintenance and quality control for TPL.

\section*{Declaration of Generative AI and AI-assisted technologies in the
writing process}
During the preparation of this manuscript, the authors used ChatGPT to help improve wording and correct English mistakes. After using this tool/service, the authors reviewed and edited the content as needed and take full responsibility for the content of the publication.

\section*{Acknowledgments}

This research has been supported by the National Science Foundation, USA under Grant No. 2434813. The experiments were carried out in part in the Materials Research Laboratory Central Research Facilities, University of Illinois.

\bibliographystyle{unsrtnat}  
\bibliography{ref}  

\begin{thebibliography}{10}
\expandafter\ifx\csname url\endcsname\relax
  \def\url#1{\texttt{#1}}\fi
\expandafter\ifx\csname urlprefix\endcsname\relax\def\urlprefix{URL }\fi
\expandafter\ifx\csname href\endcsname\relax
  \def\href#1#2{#2} \def\path#1{#1}\fi

\bibitem{harinarayana2021two}
V.~Harinarayana, Y.~Shin, Two-photon lithography for three-dimensional fabrication in micro/nanoscale regime: A comprehensive review, Optics \& Laser Technology 142 (2021) 107180.

\bibitem{vyatskikh2018additive}
A.~Vyatskikh, S.~Delalande, A.~Kudo, X.~Zhang, C.~M. Portela, J.~R. Greer, Additive manufacturing of 3d nano-architected metals, Nature communications 9~(1) (2018) 593.

\bibitem{maddox2020digitization}
S.~Maddox, M.~Afshar-Mohajer, M.~Zou, Digitization, replication, and modification of physical surfaces using two-photon lithography, Journal of Manufacturing Processes 54 (2020) 180--189.

\bibitem{dong2024filtered}
Z.~Dong, S.~Jia, C.~Shao, Filtered kriging for improved interpolation of periodic manufacturing surfaces, Journal of Manufacturing Processes 131 (2024) 1--12.

\bibitem{sun2025emerging}
J.~Sun, S.~Jia, C.~Shao, M.~R. Dawson, K.~C. Toussaint, Emerging technologies for multiphoton writing and reading of polymeric architectures for biomedical applications, Annual Review of Biomedical Engineering 27 (2025).

\bibitem{williams2018two}
G.~Williams, M.~Hunt, B.~Boehm, A.~May, M.~Taverne, D.~Ho, S.~Giblin, D.~Read, J.~Rarity, R.~Allenspach, et~al., Two-photon lithography for 3d magnetic nanostructure fabrication, Nano Research 11 (2018) 845--854.

\bibitem{jia2025end}
S.~Jia, S.~Li, J.~Sun, M.~R. Dawson, K.~C. Toussaint~Jr, C.~Shao, End-to-end part quality classification for two-photon lithography using computer vision, Manufacturing Letters 44 (2025) 1369--1377.

\bibitem{sun2024automated}
J.~Sun, A.~M. Howes, S.~Jia, J.~A. Burrow, P.~F. Felzenszwalb, M.~R. Dawson, C.~Shao, K.~C. Toussaint~Jr, Automated brightfield layerwise evaluation in three-dimensional micropatterning via two-photon polymerization, Optics Express 32~(7) (2024) 12508--12519.

\bibitem{maruo1997three}
S.~Maruo, O.~Nakamura, S.~Kawata, Three-dimensional microfabrication with two-photon-absorbed photopolymerization, Optics letters 22~(2) (1997) 132--134.

\bibitem{shunhua2023high}
Y.~Shunhua, D.~Chenliang, Z.~Dazhao, Y.~Zhenyao, L.~Yong, K.~Cuifang, L.~Xu, High-speed two-photon lithography based on femtosecond laser, Opto-Electronic Engineering 50~(3) (2023) 220133--1.

\bibitem{beermann2008two}
J.~Beermann, S.~M. Novikov, T.~S{\o}ndergaard, A.~E. Boltasseva, S.~I. Bozhevolnyi, Two-photon mapping of localized field enhancements in thin nanostrip antennas, Optics Express 16~(22) (2008) 17302--17309.

\bibitem{noronha2024titanium}
J.~Noronha, J.~Dash, J.~Rogers, M.~Leary, M.~Brandt, M.~Qian, Titanium multi-topology metamaterials with exceptional strength, Advanced Materials 36~(34) (2024) 2308715.

\bibitem{surjadi2019mechanical}
J.~U. Surjadi, L.~Gao, H.~Du, X.~Li, X.~Xiong, N.~X. Fang, Y.~Lu, Mechanical metamaterials and their engineering applications, Advanced Engineering Materials 21~(3) (2019) 1800864.

\bibitem{jalali2006raman}
B.~Jalali, V.~Raghunathan, D.~Dimitropoulos, O.~Boyraz, Raman-based silicon photonics, IEEE Journal of Selected Topics in Quantum Electronics 12~(3) (2006) 412--421.

\bibitem{langfelder2010mems}
G.~Langfelder, A.~F. Longoni, A.~Tocchio, E.~Lasalandra, Mems motion sensors based on the variations of the fringe capacitances, IEEE Sensors Journal 11~(4) (2010) 1069--1077.

\bibitem{zhou2015review}
X.~Zhou, Y.~Hou, J.~Lin, A review on the processing accuracy of two-photon polymerization, Aip Advances 5~(3) (2015).

\bibitem{wang2023two}
H.~Wang, W.~Zhang, D.~Ladika, H.~Yu, D.~Gailevi{\v{c}}ius, H.~Wang, C.-F. Pan, P.~N.~S. Nair, Y.~Ke, T.~Mori, et~al., Two-photon polymerization lithography for optics and photonics: fundamentals, materials, technologies, and applications, Advanced Functional Materials 33~(39) (2023) 2214211.

\bibitem{lee2023enhanced}
J.~Lee, S.~J. Park, S.~C. Han, P.~Prabhakaran, C.~W. Ha, Enhanced mechanical property through high-yield fabrication process with double laser scanning method in two-photon lithography, Materials \& Design 235 (2023) 112389.

\bibitem{yu2024two}
S.~Yu, Q.~Du, C.~R. Mendonca, L.~Ranno, T.~Gu, J.~Hu, Two-photon lithography for integrated photonic packaging, Light: Advanced Manufacturing 4~(4) (2024) 486--502.

\bibitem{lee2020automated}
X.~Y. Lee, S.~K. Saha, S.~Sarkar, B.~Giera, Automated detection of part quality during two-photon lithography via deep learning, Additive Manufacturing 36 (2020) 101444.

\bibitem{yang2022machine}
Y.~Yang, V.~A. Kelkar, H.~S. Rajput, A.~C.~S. Coariti, K.~C. Toussaint~Jr, C.~Shao, Machine-learning-enabled geometric compliance improvement in two-photon lithography without hardware modifications, Journal of Manufacturing Processes 76 (2022) 841--849.

\bibitem{jia2024hybrid}
S.~Jia, J.~Sun, A.~Howes, M.~R. Dawson, K.~C. Toussaint~Jr, C.~Shao, Hybrid physics-guided data-driven modeling for generalizable geometric accuracy prediction and improvement in two-photon lithography, Journal of Manufacturing Processes 110 (2024) 202--210.

\bibitem{khanafer2024condition}
K.~Khanafer, J.~Cao, H.~Kokash, Condition monitoring in additive manufacturing: A critical review of different approaches, Journal of Manufacturing and Materials Processing 8~(3) (2024) 95.

\bibitem{zhu2021metal}
K.~Zhu, J.~Y.~H. Fuh, X.~Lin, Metal-based additive manufacturing condition monitoring: A review on machine learning based approaches, IEEE/ASME Transactions on Mechatronics 27~(5) (2021) 2495--2510.

\bibitem{fang2022process}
Q.~Fang, G.~Xiong, M.~Zhou, T.~S. Tamir, C.-B. Yan, H.~Wu, Z.~Shen, F.-Y. Wang, Process monitoring, diagnosis and control of additive manufacturing, IEEE Transactions on Automation Science and Engineering 21~(1) (2022) 1041--1067.

\bibitem{liu2018improved}
J.~Liu, Y.~Hu, B.~Wu, Y.~Wang, An improved fault diagnosis approach for fdm process with acoustic emission, Journal of Manufacturing Processes 35 (2018) 570--579.

\bibitem{tlegenov2018nozzle}
Y.~Tlegenov, G.~S. Hong, W.~F. Lu, Nozzle condition monitoring in 3d printing, Robotics and Computer-Integrated Manufacturing 54 (2018) 45--55.

\bibitem{wang2024maintenance}
M.~Wang, N.~Kashaev, On the maintenance of processing stability and consistency in laser-directed energy deposition via machine learning, Journal of Manufacturing Systems 73 (2024) 126--142.

\bibitem{meng2023explainable}
Y.~Meng, K.-C. Lu, Z.~Dong, S.~Li, C.~Shao, Explainable few-shot learning for online anomaly detection in ultrasonic metal welding with varying configurations, Journal of Manufacturing Processes 107 (2023) 345--355.

\bibitem{meng2024meta}
Y.~Meng, Z.~Dong, K.-C. Lu, S.~Li, C.~Shao, Meta-learning-based domain generalization for cost-effective tool condition monitoring in ultrasonic metal welding, IEEE Transactions on Industrial Informatics (2024).

\bibitem{dong2025new}
X.~Dong, C.~Zhang, H.~Liu, D.~Wang, Y.~Chen, T.~Wang, A new cross-domain bearing fault diagnosis method with few samples under different working conditions, Journal of Manufacturing Processes 135 (2025) 359--374.

\bibitem{sun2023situ}
J.~Sun, A.~Howes, S.~Jia, J.~A. Burrow, M.~Dawson, C.~Shao, K.~C. Toussaint, In-situ monitoring and process control of two-photon lithography for tissue scaffold fabrication, in: 3D Image Acquisition and Display: Technology, Perception and Applications, Optica Publishing Group, 2023, pp. DTu3A--3.

\end{thebibliography}
\end{document}